%% file: paper.tex
\documentclass{article}

\usepackage{enumitem}
    \usepackage[preprint]{neurips_2019}

\usepackage[utf8]{inputenc} % allow utf-8 input
\usepackage[T1]{fontenc}    % use 8-bit T1 fonts
\usepackage{hyperref}       % hyperlinks
\usepackage{url}            % simple URL typesetting
\usepackage{booktabs}       % professional-quality tables
\usepackage{amsfonts}       % blackboard math symbols
\usepackage{nicefrac}       % compact symbols for 1/2, etc.
\usepackage{microtype}      % microtypography

\usepackage{amsmath,amssymb}
\usepackage{color,soul}
\input{math_commands.tex}

\newcommand{\mDelta}{\mathbf{\Delta}}
\newcommand{\bigO}{\mathcal{O}}

\usepackage[textsize=footnotesize,disable]{todonotes}
\usepackage{url}
\newcommand{\ptitle}[1]{\vspace{1mm}\noindent{\bf #1.}}
\newcommand{\ptitlenoskip}[1]{\noindent{\bf #1.}}
\newcommand{\fb}[1]{\textcolor{black}{#1}}
\newcommand{\dblp}[1]{\textcolor{black}{#1}}

\setlist[itemize]{leftmargin=*}
\setlist[enumerate]{leftmargin=*}
\setlist{nolistsep}

\title{Conditional t-SNE: Complementary t-SNE embeddings through factoring out prior information}

\author{
    Bo Kang \\
    IDLab, Ghent University\\
    \And
    Dar\'io Garc\'ia Garc\'ia \\
    Facebook AI \\
    \And
    Jefrey Lijffijt \\
    IDLab, Ghent University\\
    \And
    Ra\'ul Santos-Rodr\'iguez \\
    University of Bristol \\
    \And
    Tijl {De Bie} \\
    IDLab, Ghent University\\
}

\begin{document}

\maketitle

\begin{abstract}
Dimensionality reduction and manifold learning methods such as t-Distributed Stochastic Neighbor Embedding (t-SNE) are routinely used to map high-dimensional data into a 2-dimensional space to visualize and explore the data. However, two dimensions are typically insufficient to capture all structure in the data, the salient structure is often already known, and it is not obvious how to extract the remaining information in a similarly effective manner. To fill this gap, we introduce \emph{conditional t-SNE} (ct-SNE), a generalization of t-SNE that discounts prior information from the embedding in the form of labels. To achieve this, we propose a conditioned version of the t-SNE objective, obtaining a single, integrated, and elegant method. ct-SNE has one extra parameter over t-SNE; we investigate its effects and show how to efficiently optimize the objective. Factoring out prior knowledge allows complementary structure to be captured in the embedding, providing new insights. Qualitative and quantitative empirical results on synthetic and (large) real data show ct-SNE is effective and achieves its goal.
\end{abstract}

%===============================================================================
\section{Introduction}\label{sec:intro}
%===============================================================================
Dimensionality reduction (DR) methods can be used to create low-dimensional (typically 2-dimensional; 2-d) representations that are straightforward to visualize and subsequently explore the dominant structure of high-dimensional datasets. Non-linear DR methods are particularly powerful because they can capture complex structure even when it is spread over many dimensions. This explains the huge popularity of methods such as t-SNE \citep{maaten2008visualizing}, LargeVis \citep{tang2016visualizing}, and UMAP \citep{mcinnes2018umap}.

However, DR methods yield a single static embedding and the most prominent structure present in the data may already be known to the analyst. One may indeed construct higher-dimensional embeddings, hoping to uncover more structure, but there is no guarantee that any of the constructed dimensions is fully complementary to the prior knowledge of an analyst. Besides, the salient structure that is already known may be spread across all attributes, hence we cannot just remove the associated attributes and generally speaking it is not obvious how to visualize the remaining structure. The question arises: can we actively filter or discount prior knowledge from the embedding?

%High dimensional data contains information that has many aspects. To explore them, a common practice in exploratory data analysis is to disentangle one or a few aspects from the data using dimensionality reduction (DR) methods and study them accordingly. In order to disentangle different types of information form data, many DR methods are developed. t-distributed stochastic neighbor embedding \citep{maaten2008visualizing} is arguably one of the most popular DR methods due to its effectiveness of preserving both local and global structure in the lower dimensional representations. Once a lower dimensional representation is studied, exploring the remaining aspects of the data becomes a natural next step. However, the existing state-of-the-art DR methods such as t-SNE have no built-in mechanism to filter out the known aspects and present the unknown ones.

To this end, we introduce \emph{conditional t-SNE} (ct-SNE), a generalization of t-SNE that accounts for prior information about the data. By discounting the prior information, the embedding may focus on capturing complementary information. More concretely, it does not aim to construct an embedding that reflects all the proximities in the original data (the objective of t-SNE), but it should reflect all pairwise proximities \emph{conditioned} on whether we expect that pair to be close or not. %In effect, points that are known to be similar do not need to be embedded close to each other, creating a potential for other information and structure to be captured.

%In this manner, we can discount known factors from the low-dimensional embedding. As such, ct-SNE precisely addresses the question of how to extract complementary structure from the data using only 2-d embeddings.
%As we demonstrate in the experimental study (Sec.~\ref{sec:experiment}),
ct-SNE enables at least three new ways to obtain insight into data:
\begin{itemize}
	\item The prior knowledge may indeed be available beforehand, in which case we can straight away focus the analysis on an embedding that is more useful.
	\item Such prior knowledge may be gained during analysis, leading to an iterative data analysis process.
	\item If we observe some specific structure X in an embedding and then factor out specific information Y, then if X remains present in the embedding, we learn that X is Y complementary to Y.
\end{itemize}

Note we use the term \emph{prior knowledge}, even when this knowledge is not available a priori, but gained during the analysis. This reflects the knowledge is available prior to the embedding step.

\todo[inline,author=Jef]{We should write somewhere in the paper that even though the technical aspects of the question are addressed using ct-SNE, this does not mean we can in practice extract all structure in a sequence of 2-d embeddings. Because, in order to do so, the right combination of priors has to be used, and there is no guarantee plain iterative use would achieve that.}

%Similar to t-SNE, ct-SNE represents the proximities in the input space using lower dimensional embeddings. Unlike t-SNE, ct-SNE assumes data points are labeled according to some aspects of data known in a priori. Namely, points that are similar according to the prior have the same label, and points that are in a priori known to be different have different labels. ct-SNE further assumes proximity in the input space can be explained by the proximity in the embedding space or explained by the labels being the same, and that a large distance in the input space can be explained by a large distance in the embedding space or by the labels being different. The result is that the same-labeled points are not necessarily embedded close to each other, and differently labeled points are not necessarily embedded far from each other. Namely, the necessity of the known factors being presented in the subsequent ct-SNE embeddings is removed.

\begin{figure*}[t]
\centering
   \includegraphics[trim=0 0 0 0, clip=true, width=\textwidth]{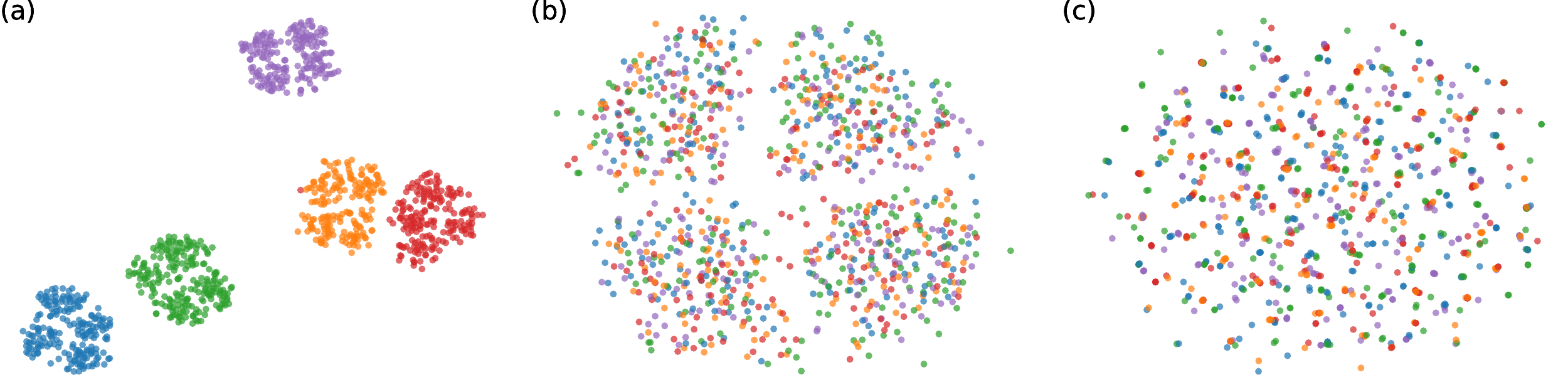}
\caption{Visualization of 2-d embeddings of synthetic data (see `example' below).\label{fig:synthetic_emb}}
\end{figure*}

\ptitle{Example} To demonstrate the idea behind ct-SNE more concretely, consider a ten-dimensional dataset with 1,000 data points. In dimension 1--4 the data points fall into five clusters (following a multi-variate Gaussian with small variance), similarly for dimensions 5--6 the points fall randomly into four clusters. Dimensions 7--10 contain Gaussian noise with larger variance. Figure~\ref{fig:synthetic_emb}a gives the t-SNE embedding. It shows five large clusters, where some can be somewhat clearly split further into smaller clusters. The large clusters correspond to those defined in dimension 1--4. Figure~\ref{fig:synthetic_emb}b is the ct-SNE embedding where we have input the five colored clusters as prior knowledge. This figure shows four clusters that are complementary to the five clusters observed in \ref{fig:synthetic_emb}a. We see they are complementary because there is no correlation between the colors and the clusters in Figure~\ref{fig:synthetic_emb}b. These four clusters are indeed those defined in dimensions 5--6. Finally, Figure~\ref{fig:synthetic_emb}c shows that after combining the labels, ct-SNE yields an embedding capturing only on the remaining noise.

%As illustrated in the example, using ct-SNE to remove the known factors from data helps to find low-dimensional representations that do not reflect the known factors. This allows structure that represents unknown aspects of the data to be further explored. As we will show in the experiment section, ct-SNE can also be used to distill information. Namely, if some structure is persistent before and after certain known factor is removed, then we known such structure is not correlated with the known factor thus worth further investigation.

The implementation of ct-SNE and code for the experiments on public data are available at \url{https://bitbucket.org/ghentdatascience/ct-sne}.

\ptitlenoskip{Contributions}
This paper makes the following contributions:
\begin{itemize}
  \item
    ct-SNE, a new DR method that searches for an embedding such that a distribution defined in terms of distances in the input space (as in t-SNE) is well-approximated by a distribution defined in terms of distances in the embedding space \emph{after conditioning on the prior knowledge} (Sec.~\ref{sec:model});
  \item
    A Barnes-Hut-Tree based optimization method to efficiently find an embedding (Sec.~\ref{sec:optimization});
  \item
    We illustrate that the concept of conditioning embeddings on prior information can be applied to other popular non-linear DR methods (mentioned in Sec.~\ref{sec:method}, with details in Appendix~\ref{app:discussion});
  \item
    Extensive qualitative and quantitative experiments on synthetic and real world datasets show ct-SNE effectively removes the known factors, enables deeper visual analysis of high-dimensional data, and that ct-SNE scales sufficiently to handle hundreds of thousands of points (Sec.~\ref{sec:experiment}).
\end{itemize}

% \todo[inline]{Bo: If we decide to make public the ct-SNE built on Laurens van der Maaten's implementation. We need to incorporate the BSD 3-clause license of the original implementation in our code package (if I understood correctly from the BSD 3-clause convention)}

%===============================================================================
\section{Method\label{sec:method}}
%===============================================================================
In this section, we first briefly recap the idea behind t-SNE and introduce the basic notation. Then, we derive ct-SNE and describe a Barnes-Hut based strategy to optimize the ct-SNE objective. Due to space limitations, we discuss in Appendix~\ref{app:discussion} how the idea of factoring out prior information can be applied to many other existing non-linear DR methods such as LargeVis and UMAP.
%-------------------------------------------------------------------------------
\subsection{Background: t-SNE}
%-------------------------------------------------------------------------------

In t-SNE, the input data set $\mX\in\sR^{n\times d}$ is taken to define a probability distribution for a categorical random variable $e$, of which the value domain is indexed by all pairs $(i,j)$ of indices $i,j\in [1..n]$ with $i\neq j$. This distribution is determined by specifying probabilities $0\leq p_{ij}\leq 1$ such that $\sum_{i\neq j} p_{ij}=1$, equal to the probability that $e=(i,j)$. For brevity, below we will speak of \emph{the distribution $\vp$} when we mean the categorical distribution with parameters $p_{ij}$.

More specifically, in t-SNE, the distribution $\vp$ is defined as follows:
\begin{align}\label{eq:high_dim_distr}
p_{ij}\triangleq& P_\vp(e=(i,j))=\frac{\exp\left(-\|\vx_i-\vx_j\|^2/2\sigma^2\right)}{\sum_{k\neq l}\exp\left(-\|\vx_k-\vx_l\|^2/2\sigma^2\right)}.
\end{align}
The goal of t-SNE is to find another embedding $\mY\in\sR^{n\times d'}$,
from which another categorical probability distribution is derived,
specified by the values $q_{ij}$ defined as follows:
\begin{align}\label{eq:low_dim_distr}
q_{ij}\triangleq& P_\vq(e=(i,j))=\frac{(1+\|\vy_i-\vy_j\|^2)^{-1}}{\sum_{k\neq l}(1+\|\vy_k-\vy_l\|^2)^{-1}}.
\end{align}
An embedding $\mY$ is deemed better if the distance between these two categorical distributions
is smaller, as quantified by the KL-divergence:
%\begin{align*}
$KL(\vp\|\vq)=\sum_{i\neq j} p_{ij}\log\left(\frac{p_{ij}}{q_{ij}}\right)$.
%\end{align*}

%-------------------------------------------------------------------------------
\subsection{Conditional t-SNE \label{sec:model}}
%-------------------------------------------------------------------------------

Let us now assume that each data point $\vx_i$ has a label $l_i$ associated, with $l_i\in[0..L]$ for all $i\in[1..n]$. Moreover, let us assume that it is known a priori that same-labeled data points are more likely to be nearby in $\mX$. Our goal is to ensure that the embedding $\mY$ does not reflect that information again. This can be achieved by minimizing the KL-divergence between the distributions $\vp$ and $\vr$ (rather than $\vq$),
where $\vr$ is the distribution derived from the embedding $\mY$ but \emph{conditioned on the prior knowledge}.

We formalize this using the following notation. The indicator variable $\delta_{ij}=1$ if $l_i=l_j$ and $\delta_{ij}=0$ if $l_i\neq l_j$, and the label matrix $\mDelta$ is defined by $[\mDelta]_{ij}=\delta_{ij}$. The probability that the random variable $e$ is equal to $(i,j)$, \emph{conditioned on} the label matrix $\mDelta$ (i.e. the prior information) is denoted as:
\begin{align*}
r_{ij}\triangleq& P_\vq(e=(i,j)|\mDelta)= \frac{P(\mDelta|e=(i,j))P_\vq(e=(i,j))}{P_\vq(\mDelta)}.
\end{align*}
In ct-SNE, this is the probability distribution that needs to be similar to $\vp$ for the embedding to be a good one.
Note that if we ensure that $P(\mDelta|e=(i,j))$ is larger when $\delta_{ij}=1$ than when $\delta_{ij}=0$, it will be less important for the embedding to ensure that $q_{ij}=P_\vq(e=(i,j))$ is large for same-labeled data points, even if $p_{ij}$ is large. I.e., \emph{for same-labeled data points}, it is less important to be embedded nearby even if they are nearby in the input representation. This is precisely the goal of ct-SNE.

To compute $P_\vq(e=(i,j)|\mDelta)$, we now investigate its different factors.
First, $P_\vq(e=(i,j))=q_{ij}$ is simply computed as in Eq.~(\ref{eq:low_dim_distr}). Second, we need to determine a suitable form for $P(\mDelta|e=(i,j))$, based on the above intuition. To do this, we assume that $\delta_{ij}$ is the sufficient statistic for $P(\mDelta|e=(i,j))$, i.e. $P(\mDelta|e=(i,j))=\alpha^{\delta_{ij}}\beta^{1-\delta_{ij}}$, where $\alpha$ and $\beta$ can be regarded as the confidence of points $\vx_i$ and $\vx_j$ being randomly picked to have the same or different labels. Let us further denote the class size of the $l$'th class as $n_l$%, and assume that the class sizes are known
. Then, for this distribution to be normalized, it must hold that:
\begin{align*}
1=& \sum_{\mDelta} P(\mDelta|e=(i,j)),\nonumber\\
=& \alpha\left(\sum_l\frac{(n-2)!}{(n_l-2)!\prod_{f\neq l}n_f!}\right)
+ \beta\left(\frac{n!}{\prod_l n_l!}-\sum_l\frac{(n-2)!}{(n_l-2)!\prod_{f\neq l}n_f!}\right),\nonumber\\
=& \frac{n!}{\prod_l n_l!}\left(\alpha\frac{\sum_l n_l(n_l-1)}{n(n-1)}+\beta\left(1-\frac{\sum_l n_l(n_l-1)}{n(n-1)}\right)\right).
\end{align*}
This yields a relation between $\alpha$ and $\beta$. It also suggests a ballpark figure for $\alpha$. Indeed, one would typically set $\alpha>\beta$.
For $\alpha=\beta$ (i.e. the lower bound for $\alpha$), they would both be equal to $\alpha=\beta=\frac{\prod_l n_l!}{n!}$, i.e. one divided by the number of possible distinct label assignments (this is of course entirely logical). Thus, in tuning $\alpha$, one could take multiples of this minimal value.

We can now also compute the marginal probability $P_\vq(\mDelta)$ as follows:
\begin{align*}
P_\vq(\mDelta)=& \sum_{i\neq j} P(\mDelta|e=(i,j))P_\vq(e=(i,j))
= \sum_{i\neq j} q_{ij}\alpha^{\delta_{ij}}\beta^{1-\delta_{ij}},\nonumber\\
=& \alpha\sum_{i\neq j:\delta_{ij}=1}q_{ij} + \beta\sum_{i\neq j:\delta_{ij}=0}q_{ij}.
\end{align*}
Given all this, one can then compute the required conditional distribution as follows:
\begin{align}\label{eq:ctsne_posterior}
r_{ij}\triangleq & P_\vq(e=(i,j)|\mDelta) = \frac{P(\mDelta|e=(i,j))P_\vq(e=(i,j))}{P_\vq(\mDelta)},\\
=&
\left\{
\begin{array}{rll}
\frac{\alpha q_{ij}}{\alpha\sum_{i\neq j:\delta_{ij}=1}q_{ij} + \beta\sum_{i\neq j:\delta_{ij}=0}q_{ij}}&\mbox{if}&\delta_{ij}=1,\\
\frac{\beta q_{ij}}{\alpha\sum_{i\neq j:\delta_{ij}=1}q_{ij} + \beta\sum_{i\neq j:\delta_{ij}=0}q_{ij}}&\mbox{if}&\delta_{ij}=0.
\end{array}
\right.\nonumber
\end{align}
It is numerically better to express this in terms of new variables
$\alpha'\triangleq\alpha\frac{n!}{\prod_l n_l!}$ and $\beta'\triangleq\beta\frac{n!}{\prod_l n_l!}$:
\begin{align*}
r_{ij}=&
\left\{
\begin{array}{rll}
\frac{\alpha' q_{ij}}{\alpha'\sum_{i\neq j:\delta_{ij}=1}q_{ij} + \beta'\sum_{i\neq j:\delta_{ij}=0}q_{ij}}&\mbox{if}&\delta_{ij}=1,\\
\frac{\beta' q_{ij}}{\alpha'\sum_{i\neq j:\delta_{ij}=1}q_{ij} + \beta'\sum_{i\neq j:\delta_{ij}=0}q_{ij}}&\mbox{if}&\delta_{ij}=0,
\end{array}
\right.
\end{align*}
where the relation between $\alpha'$ and $\beta'$ is:
\begin{align}\label{eq:relalphaprimebetaprime}
1 =& \alpha'\frac{\sum_l n_l(n_l-1)}{n(n-1)}+\beta'\left(1-\frac{\sum_l n_l(n_l-1)}{n(n-1)}\right).
\end{align}
This has the advantage of avoiding the large factorials and resulting numerical problems.
The lower bound for $\alpha'$ to be considered is now $1$ (in which case also $\beta'=1$).

Finally, computing the KL-divergence with $\vp$, yields the ct-SNE objective function to be minimized:
\begin{align}\label{eq:ct-SNE_obj}
KL(\vp\|\vr)=& \sum_{i\neq j} p_{ij}\log\left(\frac{p_{ij}}{r_{ij}}\right), \nonumber\\
=& KL(\vp\|\vq) + \sum_{i\neq j}p_{ij}\log\left(\frac{\alpha'\sum_{i\neq j:\delta_{ij}=1}q_{ij} + \beta'\sum_{i\neq j:\delta_{ij}=0}q_{ij}}
{{\alpha'}^{\delta_{ij}}{\beta'}^{1-\delta_{ij}}}\right), \nonumber\\
=& KL(\vp\|\vq) + \log\left(\alpha'\sum_{i\neq j:\delta_{ij}=1}q_{ij} + \beta'\sum_{i\neq j:\delta_{ij}=0}q_{ij}\right) \nonumber\\
& - \sum_{i\neq j:\delta_{ij}=1}p_{ij}\log(\alpha') - \sum_{i\neq j:\delta_{ij}=0}p_{ij}\log(\beta').
\end{align}
Note that the last two terms are constant w.r.t. $q_{ij}$. Moreover, it is clear that for $\alpha'=\beta'=1$, this reduces to standard t-SNE. For $\alpha'>1>\beta'$ (and related as per the Eq.~(\ref{eq:relalphaprimebetaprime})), the minimization of this KL-divergence will try to minimize $q_{ij}$ when $\delta_{ij}=1$ more strongly (as it is multiplied with the larger number $\alpha'$) than when $\delta_{ij}=0$ (when it is multiplied with the smaller number $\beta'$).

%-------------------------------------------------------------------------------
\subsection{Optimization\label{sec:optimization}}
%-------------------------------------------------------------------------------

The objective function (Eq.~(\ref{eq:ct-SNE_obj})) is non-convex w.r.t the embedding $\mY$. Even so, we find that optimizing the objective function using gradient descent with random restarts works well in practice. The gradient of the objective function w.r.t. the embedding of a point $\vy_i$ reads: \footnote{A detailed derivation of the gradient computation can be found in Appendix~\ref{app:grad_der}.}
\begin{align*}
  & \nabla_{\vy_i}KL(\vp\|\vr) = 4\left(F_{\text{attr}} + F_{\text{rep}}\right),\nonumber \\
  & =4\sum_j\left(p_{ij}q_{ij}Z(\vy_i - \vy_j) - \frac{\delta_{ij}\alpha' + (1-\delta_{ij})\beta'}{O}\cdot q_{ij}^2Z(\vy_i - \vy_j)\right).
\end{align*}
where $Z = \sum_{k\neq l} (1+\|\vy_k-\vy_l\|^2)^{-1}$ and $O = \alpha' \sum_{i\neq j: \delta_{kl} = 1}q_{kl} + \beta'\sum_{i\neq j: \delta_{kl} = 0}q_{kl}$.
The gradient can be decomposed in attraction and repelling forces between points in the embedding space. Thus the underlying problem of ct-SNE, just like many other force-based embedding methods, is related to the classic $n$-body problem in physics\footnote{\url{https://en.wikipedia.org/wiki/N-body_problem\#Other_n-body_problems}}, which has also been studied in the recent machine learning literature \citep{gray2001n, ram2009linear}. The general goal of the $n$-body problem is to find a constellation of $n$ objects such that equilibrium is achieved according to a certain measure (e.g., forces, energy). In the problem setting of ct-SNE, both the pairwise distances between points and the label information affect the attraction and repelling forces. Particularly, the label information strengthens the repelling force (assume $\alpha' > 1 > \beta' > 0$) between two points if they have the same label and weakens the repelling force if two points have different labels. This is desirable behavior because we do not want to reflect the known label information in the resulted embeddings.

Evaluating the gradient has complexity $\bigO(n^2)$, which makes the computation (both time and memory cost) infeasible when $n$ is large (e.g., $n> 100k$). As an approximation of the gradient computation, we adapt the tree-based approximation strategy described by \citet{van2014accelerating}. To efficiently model the proximity in high-dimensional space (Eq.~(\ref{eq:high_dim_distr})) we use a vantage-point tree-based algorithm (which exploits the fast diminishing property of the Gaussian distribution). To approximate the low-dimensional proximity (Eq.~(\ref{eq:ctsne_posterior})) we modify the Barnes-Hut algorithm to incorporate the label information. The basic idea of the Barnes-Hut algorithm is to organize the points in the embedding space using a kd-tree (which for 2-d embeddings is equivalent to a quad tree). Each node of the tree corresponds to a cell (dissection) in the embedding space. If a target point $\vy_i$ is far away from all the points in a given cell, then the interaction between the target point and the points within the cell can be summarized by the interaction between $\vy_i$ and the cell's center of mass $\vy_{\text{cell}}$ that is computed while constructing the kd-tree. More specifically, the summarization happens when $r_{\text{cell}}/\|\vy_i - \vy_{\text{cell}}\|^2 < \theta$, where $r_{\text{cell}}$ is the radius of the cell, while $\theta$ controls the strength of summarization, i.e. the approximation strength. The summarized repelling force in t-SNE reads $F_{\text{rep}} = -n_{\text{cell}}q^2_{i,\text{cell}}Z(\vy_i - \vy_{cell})$, where $n_{\text{cell}}$ is the number of data points in that cell.

In the ct-SNE approximation, we had to overcome an additional complication though: we also need to summarize the label information for the points in a cell when the summarization happens. This can be done by maintaining a histogram in each cell, and counting the numbers of data points with different labels that fall into that cell. Then the repelling force of a target point $\vy_i$ can be weighted proportional to the number of points that have same (different) label(s) within the cell. Namely:
\begin{align*}
F^{\text{approx.}}_{\text{rep}} = -\frac{\alpha'n_{\text{cell},l = l_i} + \beta'(n_{\text{cell}} - n_{\text{cell},l = l_i} )}{O}q^2_{i,\text{cell}}Z(\vy_i - \vy_{cell}),
\end{align*}
where $n_{\text{cell},l = l_i}$ is the number of data points in a cell that has the same label as point $\vy_i$.

As both tree-based approximation schemes have complexity $\bigO\left(n\log n\right)$, counting the label will add an extra multiplicative constant $L$, equal to the number of label values in the prior information. Thus the final complexity of approximated ct-SNE is $\bigO\left(L\cdot n\log n \right)$.

%===============================================================================
\section{Experiments}\label{sec:experiment}
%===============================================================================
% We conduct empirical experiments to investigate the following questions: \textbf{Q1} Does ct-SNE work as expected in finding complementary structure? (Sections~\ref{sec:experiment_settings}--\ref{sec:exp_facebook}) \textbf{Q2} How should $\alpha$ or $\beta$ be chosen? (Section~\ref{sec:exp_param}) \textbf{Q3} Could this effect be achieved also by using (a combination) of other methods? (Section~\ref{sec:exp_baseline}) \textbf{Q4} How well does ct-SNE scale? (Section~\ref{sec:exp_runtime}).

The experiments investigate 4 questions: \textbf{Q1} Does ct-SNE work as expected in finding complementary structure? \textbf{Q2} How should $\alpha$ (or equivalently, $\beta$) be chosen? \textbf{Q3} Could ct-SNE's goal be achieved also by using (a combination of) other methods? \textbf{Q4} How well does ct-SNE scale? Two case studies addressing \textbf{Q1} are presented in Sections~\ref{sec:experiment_settings}--\ref{sec:exp_facebook}. Two more case studies addressing \textbf{Q1} as well as the experiments addressing
\textbf{Q2}--\textbf{Q4} are summarized in Sec.~\ref{sec:extra_exp}, and described in detail in Appendix~\ref{app:experiments}.

%-------------------------------------------------------------------------------
\subsection{Datasets used, and experimental settings}\label{sec:experiment_settings}
%-------------------------------------------------------------------------------

%%-------------------------------------------------------------------------------
%\subsection{Datasets}\label{sec:datasets}
%%-------------------------------------------------------------------------------
The first dataset used in the main paper is a {\bf Synthetic dataset} consisting of 1000 ten-dimensional data points, as explained in Section~\ref{sec:intro}.
The second in the main paper is a {\bf Facebook dataset} \fb{consisting of $128$-dimensional embedding of a de-identified random sample of $500k$ Facebook users in the US. This embedding is generated based purely on the list of pages and groups that the users follow, as part of an effort to improve the quality of several recommendation systems at Facebook.}

To study \textbf{Q1}, both qualitative and quantitative experiments were performed on the synthetic dataset. On the Facebook dataset we only conducted a qualitative evaluation (given the lack of ground truth).

\ptitle{Qualitative experiment} We qualitatively evaluate the effectiveness of ct-SNE through visualizations. More specifically, we compare the t-SNE visualization of a dataset with the ct-SNE visualization that has taken into account certain prior information that is visually identifiable from the t-SNE embedding. Thus by inspecting the presence of the prior information in the ct-SNE embedding and comparing to the t-SNE embedding, we can evaluate whether the prior information is removed. Conversely, we test whether information present in the ct-SNE embedding could have been identified from the t-SNE embedding to verify whether it indeed contains complementary information.

To select the prior information, we first visualize the t-SNE embedding and manually select points that are clustered in the visualization. Then we perform a \emph{feature ranking} procedure to identify the features that separate the selected points from the rest. This is done by fitting a linear classifier (logistic regression) on the selected cluster against all other data points. By inspecting the weights of the classifier, we can identify the feature that contributes the most to the classifier. Repeating this \emph{feature ranking} procedure for other clusters, we aim to find a feature that correlates with the majority of the clusters in the t-SNE visualization. This feature is then treated as prior information and provided as input to ct-SNE. In the reported experiments, the most prominent feature was always categorical, so all points with the same value were treated as being in a cluster to define the prior. We apply exact ct-SNE on Synthetic and approximated ct-SNE ($\theta=0.5$) on the Facebook dataset.

We also evaluated whether ct-SNE can continuously provide new insights, by repeatedly applying the cluster selection and feature ranking procedure on ct-SNE embeddings.

\ptitle{Quantitative experiment} In this experiment, we quantify the presence of certain prior information in a ct-SNE embedding that also take the same prior information as input. For example, if ct-SNE encodes the prior information using labels, the strong presence of certain prior information is equivalent to the high homogeneity of the encoded labels in the embedding, i.e., points that are close to each other in the embedding often have the same label. To quantify such homogeneity, we developed a measure termed \emph{normalized Laplacian score} defined as follows. Given an embedding $\mY$ and parameter $k$, we denote $\mA_k$ as the adjacency matrix of the k-nearest graph computed from the embedding. Then, the Laplacian matrix of the kNN graph has the form $\mL_k = \mA_k - \mD_k$ where $\mD_k = \text{diag}(\text{sum}(\mA_k, 1))$. We further normalize the Laplacian matrix ($\mD_k^{-1/2}\mL_k\mD_k^{-1/2}$) to obtain a score that is insensitive to the node degrees. Given a label vector $\vf$ with $L$ values where each label $l$ has $n_l$ points, and denote the one-hot encoding for each label $l$ as $\vf_l$, then the normalized Laplacian score can be computed as:
\begin{equation}\label{eq:laplacian_score}
\sum_{l \in [0..L]} \frac{n_l}{n} \vf_l'\mD_k^{-1/2}\mL_k\mD_k^{-1/2}\vf_l.
\end{equation}
This score is essentially the pairwise difference (in terms of labels) between the data points that are connected according to the kNN graph. If a label is locally consistent (homogeneous) in an embedding, the feature difference among the kNN graph neighborhood is small, which results in a small Laplacian score. Conversely, a less homogeneous label over the kNN graph would have a large Laplacian score. Thus, if ct-SNE removes certain prior information from its embedding, then the embedding should have a large Laplacian score on the labels that encode the prior information.

\begin{figure*}[!t]
	\centering
	\includegraphics[trim=0 0 0 0, clip=true, width=\textwidth]{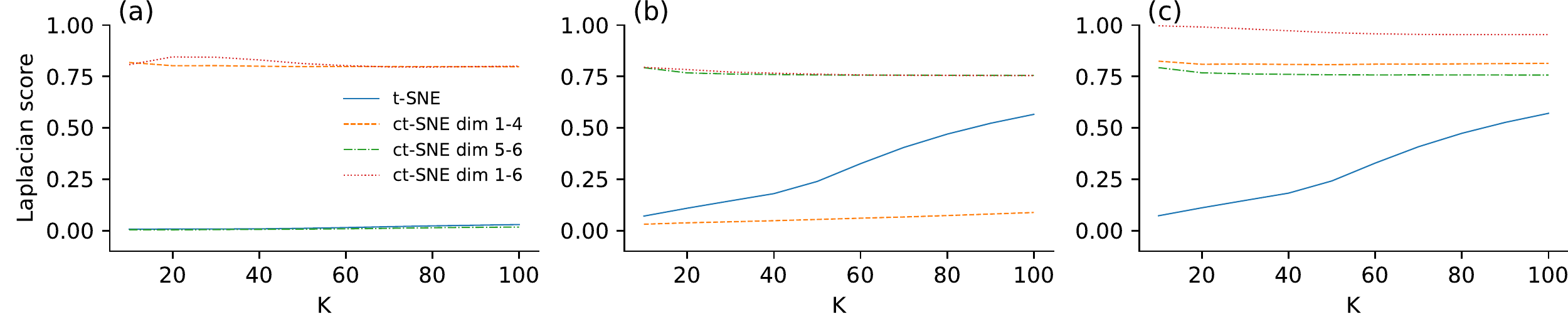}
	\vskip -0.2cm
	\caption{The homogeneity of cluster labels in t-SNE and several ct-SNE embeddings of the synthetic dataset for $k$ (a parameter of the Laplacian score) ranging from $10$ to $100$, for the three label sets: (a) $\vf_{1-4}$, (b) $\vf_{5-6}$, and (c) $\vf_{1-6}$. Colored lines give the scores for different embeddings: t-SNE (blue), ct-SNE with prior $\vf_{1-4}$ (orange), ct-SNE with prior $\vf_{5-6}$ (green), and ct-SNE with prior $\vf_{1-6}$ (red).
		%(a) The labels $\vf_{1-4}$ are less homogeneous (higher Laplacian score) in the ct-SNE embedding with prior $\vf_{1-4}$ and $\vf_{1-6}$ than in the t-SNE embedding, indicating ct-SNE effectively discounted the prior. Meanwhile both t-SNE and ct-SNE with prior $\vf_{5-6}$ clearly pick up the cluster in $\vf_{1-4}$, as indicated by the very low Laplacian score. (b)(c) ct-SNE similarly removes the prior information effectively for labels $\vf_{5-6}$ and $\vf_{1-6}$, respectively, given the associated priors. Notably, in (b) the cluster labels $\vf_{5-6}$ are most clear in the ct-SNE embedding with prior $\vf_{1-4}$ while the t-SNE embedding places smaller sets of points from these clusters (about 250 points each) close to each other (low score at $k \leq 20$), but the full clusters are spread out over the embedding.
		% (a) The labels $\vf_{1-4}$ have high homogeneity (small Laplacian score) in t-SNE embedding (blue line) as well as ct-SNE embedding with prior $\vf_{5-6}$ (green line). On the other hand, labels $\vf_{1-4}$ has low homogeneity (high Laplacian score) in ct-SNE embeddings with prior $\vf_{1-4}$ (orange line) and $\vf_{1-6}$ (red line). (b) The cluster labels $\vf_{5-6}$ have low homogeneity for t-SNE embedding and ct-SNE with prior $\vf_{1-4}$. The label $\vf_{1-4}$ also has low homogeneity in the rest two ct-SNE embeddings. (c) The cluster labels $\vf_{1-6}$ has high homogeneity in the t-SNE embedding only.
		\label{fig:synthetic_homogeneity}}
	\vskip -0.4cm
\end{figure*}

%-------------------------------------------------------------------------------
\subsection{Case study: Synthetic dataset}\label{sec:case_synthetic}
%-------------------------------------------------------------------------------

\ptitle{Qualitative experiment} The t-SNE visualization of the synthetic dataset shows five large clusters (Fig.~\ref{fig:synthetic_emb}a). Feature ranking (Sec.~\ref{sec:experiment_settings}) shows these clusters correspond to the clustering in dimensions $1$-$4$ of the data. Taking the cluster labels in dimensions $1$-$4$ ($\vf_{1-4}$) as prior, ct-SNE gives a different visualization (Fig.~\ref{fig:synthetic_emb}b). The feature ranking further shows the ct-SNE embedding indeed reveals the clusters in the dimension 5-6 of the data. We further combine the labels $\vf_{1-4}$ and $\vf_{5-6}$ by assigning a new label to each combinations of the label in $\vf_{1-4}$ and $\vf_{5-6}$, denoted as $\vf_{1-6}$. ct-SNE with $\vf_{1-6}$ yields an embedding based only on the remaining noise (Fig.~\ref{fig:synthetic_emb}c).

% \ptitle{Qualitative experiment} The t-SNE visualization of the synthetic dataset shows five large clusters (color coded in Fig.~\ref{fig:synthetic_emb}a). Feature ranking (as described in Sec.~\ref{sec:experiment_settings}) shows these clusters correspond to the clustering in dimension $1$-$4$ of the data. Taking the cluster labels in dimension $1$-$4$ ($\vf_{1-4}$) as prior, ct-SNE gives a visualization with four mix-colored clusters (Fig.~\ref{fig:synthetic_emb}b). This indicates the ct-SNE embedding is different from the t-SNE embedding. The feature ranking further shows the ct-SNE embedding indeed reveals the clusters in the dimension 5-6 of the data. We further combine the labels $\vf_{1-4}$ and $\vf_{5-6}$ by assigning a new label to each combinations of the label in $\vf_{1-4}$ and $\vf_{5-6}$, denoted as $\vf_{1-6}$. ct-SNE with $\vf_{1-6} $ as prior again gives a different embedding with no cluster structure shown (Fig.~\ref{fig:synthetic_emb}c), namely, the remaining Gaussian noise is picked up.

\ptitle{Quantitative experiment} We computed the normalized Laplacian scores (Eq.~(\ref{eq:laplacian_score})) of the t-SNE and several ct-SNE embeddings. Subfigures in Fig.~\ref{fig:synthetic_homogeneity}a--c give the Laplacian score for three label sets: $\vf_{1-4}$, $\vf_{5-6}$, and $\vf_{1-6}$. Fig.~\ref{fig:synthetic_homogeneity}a shows that labels $\vf_{1-4}$ are less homogeneous (higher Laplacian score) in the ct-SNE embeddings with prior $\vf_{1-4}$ and $\vf_{1-6}$ than in the t-SNE embedding, indicating that ct-SNE effectively discounted the prior from the embeddings. Both the t-SNE embedding and ct-SNE with prior $\vf_{5-6}$ clearly pick up the cluster in $\vf_{1-4}$, as indicated by the very low Laplacian score. Similarly, Figures~\ref{fig:synthetic_homogeneity}b,c show that ct-SNE removes the prior information effectively for labels $\vf_{5-6}$ and $\vf_{1-6}$, respectively, given the associated priors.

% \ptitle{Quantitative experiment} We computed the normalized Laplacian scores (Eq.~\ref{eq:laplacian_score}) of the t-SNE and several ct-SNE embeddings  against parameter $k$ (as in kNN graph) varying from $10$ to $100$ with step size $10$. Subfigures (a)--(c) give the Laplacian score for three label sets: $\vf_{1-4}$, $\vf_{5-6}$, and $\vf_{1-6}$. Figure~\ref{fig:synthetic_homogeneity}a shows labels $\vf_{1-4}$ are less homogeneous (higher Laplacian score) in the ct-SNE embeddings with prior $\vf_{1-4}$ and $\vf_{1-6}$ than in the t-SNE embedding, as they should be. This indicates ct-SNE effectively discounted the prior (cluster labels $\vf_{1-4}$ or $\vf_{1-6}$, respectively) from the embeddings. Both the t-SNE embedding and ct-SNE with prior $\vf_{5-6}$ clearly pick up the cluster in $\vf_{1-4}$, as indicated by the very low Laplacian score.

% Similarly, Figure~\ref{fig:synthetic_homogeneity}b,c show ct-SNE removes the prior information effectively for labels $\vf_{5-6}$ and $\vf_{1-6}$, respectively, given the associated priors. Notably, in Figure~\ref{fig:synthetic_homogeneity}b it can been seen that the cluster labels $\vf_{5-6}$ are most clear in the ct-SNE embedding with prior $\vf_{1-w4}$. The clusters each contain 250 points, and the t-SNE embedding (blue line) places smaller sets of points from these clusters close to each other (low score at $K \leq 20$), but the full clusters are spread out over the embedding.

\begin{figure*}[!t]
	\centering
	\includegraphics[width=0.8\textwidth]{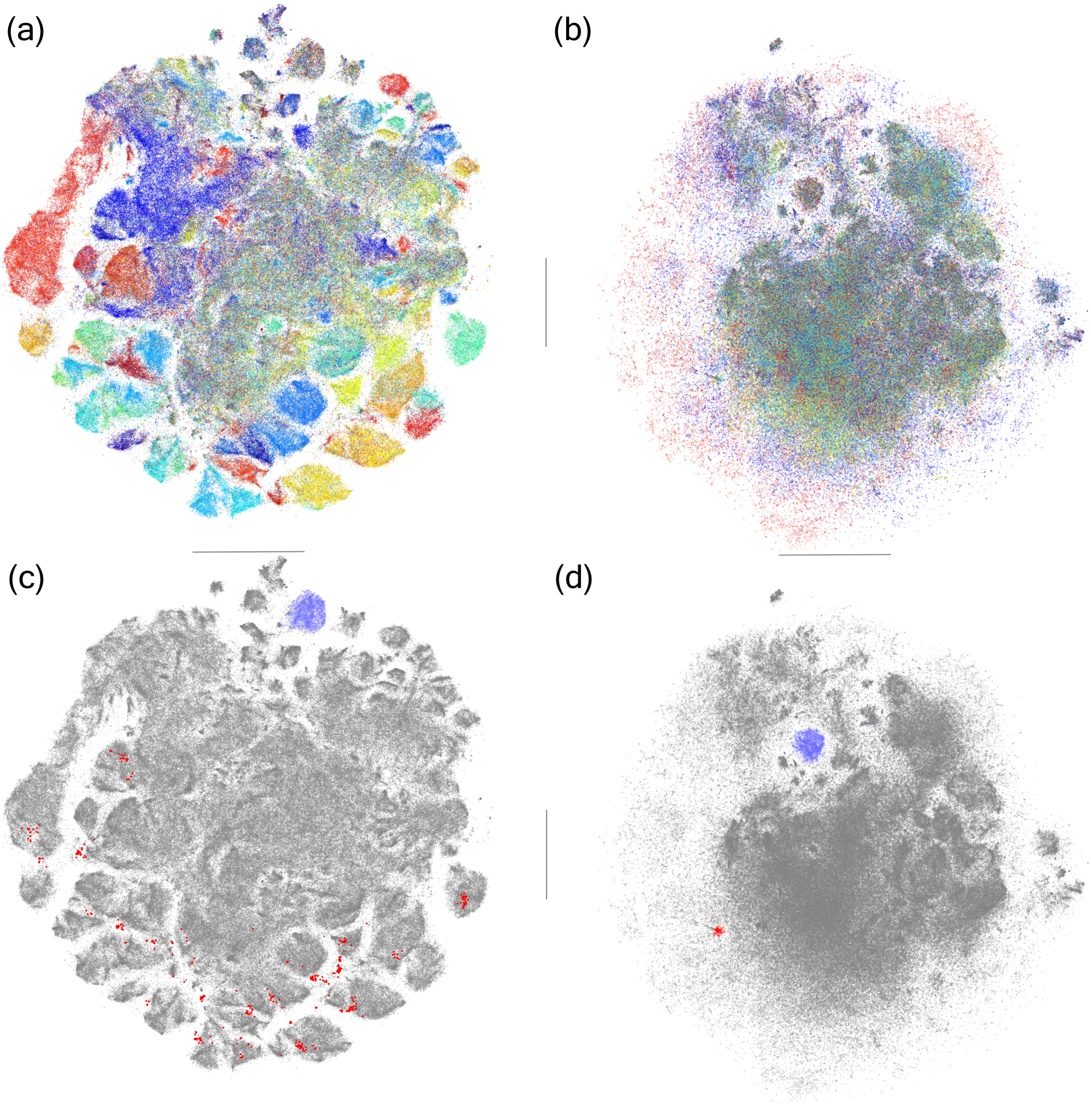}
	\vskip -0.4cm
	\caption{\fb{Visualization of 2-d embeddings of the Facebook dataset. Left column: t-SNE embedding, right column: ct-SNE embedding with region as prior. The two rows show identical embeddings but with different cluster markings (colors). See Section~\ref{sec:exp_facebook} for further info.%(a) t-SNE shows a clustering according to geographical regions (points are colored according to the regions). (b) ct-SNE shows a different clustering, with geographical information removed. (d) Highlighted in red: a newly emerged visual cluster in ct-SNE embedding spread over in the t-SNE embedding (c), corresponding to users interested in horse riding. (d) Highlighted in blue: a cluster stood-out in the ct-SNE embedding also exists in the t-SNE embedding (c), corresponding to users with an interest in Indian culture. These are merely two out of many clusters that we found to exhibit a much more informative, interest-centric structure than the t-SNE projection.}
	\label{fig:facebook_emb}}}
	\vskip -0.4cm
\end{figure*}

%-------------------------------------------------------------------------------
\subsection{Case study: Facebook dataset}\label{sec:exp_facebook}
%-------------------------------------------------------------------------------

\ptitle{Qualitative experiment}
\fb{Applying t-SNE on the Facebook dataset gives a visualization with many visually salient clusters (Fig.~\ref{fig:facebook_emb}a). Computing the feature ranking for classification of selected clusters shows that the geography (i.e., the states) contributes to the embedding the most. This is further confirmed by coloring the data points according to the geographical region in the visualization as shown in Fig.~\ref{fig:facebook_emb}a: most of the clusters are indeed quite homogeneous with respect to geography.}

\fb{To understand the effect of an embedding like this in a downstream recommendation system, an analyst would want to know what type of user interests the embedding is capturing. For this, the regional clusters are not very informative. To alleviate that we can encode the region as prior for ct-SNE so that other interesting structures can emerge in the visualization. Using the same coloring scheme, ct-SNE shows a cluster with large mass that consists of users from different states (Fig.~\ref{fig:facebook_emb}b). There are also a few small clusters with mixed color scattered on the periphery of the visualization. The visualization indicates that geographical information is mostly removed in the ct-SNE embedding. This is further confirmed by selecting clusters (highlighted in red color) in ct-SNE embedding (Fig.~\ref{fig:facebook_emb}d) and highlighting the same set of points in the t-SNE embedding (Fig.~\ref{fig:facebook_emb}c). The cluster highlighted in the ct-SNE embedding spreads over the t-SNE embedding, indicating these users are not geographically similar. Indeed, feature ranking (Sec.~\ref{sec:experiment_settings}) indicates that the selected group of users (Fig.~\ref{fig:facebook_emb}d) share an interest in horse riding: they tend to follow several pages related to that topic. Interestingly, we noticed that some of the clusters in the ct-SNE embedding are also clustered in the t-SNE embedding.  These clusters are indeed not homogeneous in terms of the geographical regions. For example, the cluster highlighted in blue in the ct-SNE embedding (Fig.~\ref{fig:facebook_emb}d) also exists in the t-SNE embedding (Fig.~\ref{fig:facebook_emb}c). Using feature ranking as above we found that these clusters are not homogeneous in terms of geography, but in terms of users' interest in Indian culture. While these clusters can thus also be seen in the t-SNE embedding, ct-SNE removes the irrelevant (region) cluster structure, such that those other clusters become more salient and easy to observe.}

%-------------------------------------------------------------------------------
\subsection{Summary of additional experimental findings}\label{sec:extra_exp}
%-------------------------------------------------------------------------------
Two other case studies (App.~\ref{app:exp_uci_adult}--\ref{app:exp_dblp}) on the UCI adult dataset \citep{dua2017} and a DBLP citation network dataset \citep{tang2008arnetminer} confirm the ability of ct-SNE visualizations to reveal insightful clusters after conditioning on prior information that dominates the t-SNE visualizations (\textbf{Q1}).
In Appendix~\ref{app:exp_param} we also analyzed the sensitivity of the ct-SNE embedding with respect to the hyperparameter $\alpha'$ (or $\beta'$) (\textbf{Q2}). By varying the hyperparameter, we found ct-SNE yields low-dimensional embeddings that better approximate the original data than t-SNE (i.e., smaller KL-divergence).
The analysis also shows that using a small $\beta'$ (e.g., $\beta' = 0.01$) is a good rule of thumb when using ct-SNE for visualization. To answer \textbf{Q3}, we compared ct-SNE to two non-trivial baselines that remove the known factors from the high-dimensional data using either an adversarial auto-encoder (AAE \citep{makhzani2015adversarial}) or canonical correlation analysis (CCA \citep{hotelling1936relations}) and then apply t-SNE for visualization (App.~\ref{app:exp_baseline}). We show that these baselines are either difficult to tune (AAE-based baseline) or have limited applicability (CCA-based baseline), while ct-SNE has essentially only one parameter to tune, and does not suffer from the limitations of the CCA baseline. Finally, we conducted a runtime experiment (App.~\ref{app:exp_runtime}) showing that the approximated ct-SNE can efficiently embed large, high-dimensional data, without substantial quality loss (\textbf{Q4}).

%===============================================================================
\section{Related Work}
%===============================================================================
Many dimensionality reduction methods have been proposed in the literature. Arguably, $n$-body problem based methods such as MDS \citep{torgerson1952multidimensional}, Isomap \citep{tenenbaum2000global}, t-SNE \citep{maaten2008visualizing}, LargeVis \citep{tang2016visualizing}, and UMAP \citep{mcinnes2018umap} appear to be the most popular ones. These methods typically have three components: (1) a proximity measure in the input space, (2) a proximity measure in the embedding space, (3) a loss function comparing the proximity between data points in the embedding space with the proximity in the input space. ct-SNE belongs to this class of DR methods. It accepts both high-dimensional data and priors about the data as inputs, and searches for low-dimensional embeddings while discounting structure in the input data specified as prior knowledge.

As a core component of ct-SNE is the prior information specified by the user, it can be considered an interactive DR method. Closely related to ct-SNE, there is a group of interactive DR methods that adjust the algorithms according to a user's inputs \citep[e.g.,][]{kang2016subjectively, puolamaki2018interactive, diaz2014interactive, alipanahi2011, barshan2011, paurat2013}. These methods contrast with ct-SNE in that the user feedback must be obeyed in the output embedding, while for ct-SNE the prior knowledge defined by the user guides what is irrelevant to the user.\footnote{For an extended discussion about the related work, please refer to Appendix~\ref{app:related_work}.}

%===============================================================================
\section{Conclusion\label{sec:conclusion}}
%===============================================================================
We introduce conditional t-SNE to efficiently discover \emph{new} insights from high-dimensional data. ct-SNE finds the lower dimensional representation of the data in a non-linear fashion while removing the known factors. Extensive case studies on both synthetic and real-world datasets demonstrate that ct-SNE can effectively remove known factors from low-dimensional representations, allowing new structure to emerge and providing new insights to the analyst. A tree-based optimization method allows ct-SNE to scale to a high dimensional dataset with hundreds of thousands of data points.

% As the future work, developing a more flexible way (e.g., continuous labels) of encoding the prior is certainly worth further investigation. Another interesting line of future work is to investigate the effect of different hyperparameter settings of ct-SNE. For example, if we set $0<\alpha'< 1< \beta'$, ct-SNE will instead of removing but finding low-dimensional representations that confirm the prior. This is a desirable feature in confirmatory data analysis. Finally, generalizing the conditioning idea to other n-body problem-based methods is also worth exploring.

\todo[inline,author=Jef]{We could remove future work if in need of space.}

% Future work:
% \begin{enumerate}
%   \item
%     Richer language for label encoding .
%   \item
%     Automatic cluster selection interpretation using cluster dispersion measure.
%   \item
%     Use ct-SNE also in confirmatory data analysis (i.e., set $0<\alpha< 1< \beta$)
%   \item
%     Apply ct-SNE conditioning idea in other n-body based methods.
%   \item
%     N-body problem optimization in general,
%   \item
%     Privacy preserving data analysis.
%   \item
%     Remove the input space independence assumption.
% \end{enumerate}

%-------------------------------------------------------------------------------
\section{Acknowledgement}
%-------------------------------------------------------------------------------
The research leading to these results has received funding from the European Research Council under the European Union's Seventh Framework Programme (FP7/2007-2013) / ERC Grant Agreement no. 615517, from the FWO (project no. G091017N, G0F9816N), from the European Union's Horizon 2020 research and innovation programme and the FWO under the Marie Sklodowska-Curie Grant Agreement no. 665501, and from the EPSRC (SPHERE EP/R005273/1). We thank Laurens van der Maaten for helpful discussions.

\bibliography{paper}
\bibliographystyle{icml2019}
\newpage

\appendix

%===============================================================================
% \section{Code and data availability}\label{app:availability}
%===============================================================================
% The implementation of ct-SNE and the code for experiments on public data are available (currently anonymized) at \url{https://tinyurl.com/yc974v9p}. After the review process, the code will be made publicly available on Bitbucket and will be free for any use.

%===============================================================================
\section{Detailed derivation of the gradient of the ct-SNE objective function}\label{app:grad_der}
%===============================================================================
Here we derive in detail the gradient of the ct-SNE objective function. Denote the euclidean distance between points as $d_{ij} \triangleq \|\vy_i - \vy_j\|_2$. The derivative of $d_{ij}$ with respect to embedding $\vy_i$ reads:
\begin{align*}
  \nabla_{\vy_i}d_{ij} = \frac{\vy_i - \vy_j}{d_{ij}}.
\end{align*}
Denote the cost (KL-divergence) by $C$:
\begin{align*}
  C & = KL(\vp\|\vr) \\
    & = C_1 + C_2 - \sum_{k\neq l: \delta_{kl}=1}p_{kl}\log(\alpha') - \sum_{i\neq j: \delta=0}p_{kl}\log(\beta'),
\end{align*}
where
\begin{align*}
  C_1 = KL(\vp \| \vq),
\end{align*}
and
\begin{align*}
  C_2 = \log\left(\alpha' \sum_{k\neq l: \delta_{kl} = 1}q_{kl} + \beta'\sum_{k\neq l: \delta_{kl} = 0}q_{kl}\right).
\end{align*}
Following the derivation from tSNE paper, the derivative of $C_1$ with respect to $\vy_i$ reads:
\begin{align*}
  \nabla_{\vy_i}C_1 = 4\sum_j(p_{ij} - q_{ij})(1 + \|\vy_i - \vy_j\|^2)^{-1}(\vy_i - \vy_j).
\end{align*}
To compute the derivative of $C_2$ with respect to $\vy_i$, we first have:
\begin{align*}
    \nabla_{\vy_i}C_2 = 2\sum_j\frac{\partial C_2}{\partial d_{ij}} \cdot \frac{\vy_i-\vy_j}{d_{ij}}.
\end{align*}
Denote $O = \alpha' \sum_{i\neq j: \delta_{kl} = 1}q_{kl} + \beta'\sum_{i\neq j: \delta_{kl} = 0}q_{kl}$The derivative of $C_2$ with respect to $d_{ij}$ can be computed as:
\begin{align*}\label{eq:derivative_c2}
  &\frac{\partial C_2}{\partial d_{ij}} = \frac{1}{O}\frac{\partial}{\partial d_{ij}}\left(\alpha' \sum_{k\neq l: \delta_{kl} = 1}q_{kl} + \beta'\sum_{k\neq l: \delta_{kl} = 0}q_{kl} \right) \nonumber \\
  &= \frac{1}{O}\left(\alpha' \sum_{k\neq l: \delta_{kl} = 1}\frac{\partial q_{kl}}{\partial d_{ij}} + \beta'\sum_{k\neq l: \delta_{kl} = 0}\frac{\partial q_{kl}}{\partial d_{ij}} \right) \nonumber \\
  &= \frac{1}{O}\Bigg(\alpha'\bigg(-2\delta_{ij}q_{ij}(1+d_{ij}^2)^{-1}d_{ij} +2\sum_{k\neq l:\delta_{kl}=1} q_{kl}q_{ij}(1+d_{ij}^2)^{-1}d_{ij}\bigg) \nonumber\\
  &+ \beta'\bigg(-2(1-\delta_{ij})q_{ij}(1+d_{ij}^2)^{-1}d_{ij} +2\sum_{k\neq l:\delta_{kl}=0} q_{kl}q_{ij}(1+d_{ij}^2)^{-1}d_{ij}\bigg)\Bigg) \nonumber \\
  &= \frac{1}{O}\Bigg(2\alpha'\bigg(-\delta_{ij} +\sum_{k\neq l:\delta_{kl}=1} q_{kl}\bigg)q_{ij}(1+d_{ij}^2)^{-1}d_{ij} \nonumber\\
  &+ 2\beta'\bigg(-(1-\delta_{ij}) +\sum_{k\neq l:\delta_{kl}= 0} q_{kl}\bigg)q_{ij}(1+d_{ij}^2)^{-1}d_{ij}\Bigg) \nonumber\\
  &=2\left(1-\frac{\delta_{ij}\alpha' + (1-\delta_{ij})\beta'}{O}\right)\cdot q_{ij}(1+d_{ij}^2)^{-1} d_{ij}.
\end{align*}
Thus we have derivative of $C_2$ with respect to $\vy_i$
\begin{align*}
    \nabla_{\vy_i}C_2 = 4\sum_j\left(1-\frac{\delta_{ij}\alpha' + (1-\delta_{ij})\beta'}{O}\right)\cdot q_{ij}(1+d_{ij}^2)^{-1} \cdot (\vy_i-\vy_j).
\end{align*}
Finally, we have derivative:
\begin{align*}
  &\nabla_{\vy_i}C = \nabla_{\vy_i}C_1 + \nabla_{\vy_i}C_2\nonumber\\
  &= 4\sum_j\left(p_{ij} - \frac{\delta_{ij}\alpha' + (1-\delta_{ij})\beta'}{O}\cdot q_{ij}\right)\cdot(1+\|\vy_i-\vy_j\|^2)^{-1}(\vy_i - \vy_j).
\end{align*}

%===============================================================================
\section{On generalizing the idea of ct-SNE\label{app:discussion}}
%===============================================================================

The idea of removing known factors from low-dimensional representations can be generalized to other $n$-body problem based DR methods. Oftentimes, the gradient of the $n$-body problem based methods can be viewed as a summation of attraction forces and repelling forces. Removing a known factor thus amounts to re-weighting the attracting and repelling forces such that points that have the same label repel each other and points with different labels attract each other.
For example, LargeVis \citep{tang2016visualizing} differs from t-SNE by modeling input space proximity using random KNN graph. Thus we can use the same conditioning idea as in ct-SNE to remove the known factors in LargeVis. However, for Uniform Manifold Approximation and Projection (UMAP) \cite{mcinnes2018umap}, conditioning is not readily applicable. In contrast to t-SNE, UMAP uses fuzzy sets to model the proximity in both input space and embedding space. Then the cross entropy between two fuzzy sets serves as loss function to compare the modeled proximity between input space and the embedding space. In the UMAP setting, it is not straightforward to condition the lower dimensional proximity model on the prior. But we can still directly re-weight the repelling forces: for data points with the same label, the pushing effect is strengthened by $\alpha$; for samples with different labels, the pushing effect is weakened by multiplying with $\beta$, with assumption $\alpha > 1 > \beta > 0$. However, without proper conditioning, parameter $\alpha$ and $\beta$ loose their probabilistic interpretation and along with it their one-to-one correspondence (as in ct-SNE), thus both parameters $\alpha$ and $\beta$ need to be set.

%===============================================================================
\section{Extended experiments\label{app:experiments}}
%===============================================================================

%-------------------------------------------------------------------------------
\subsection{Datasets}\label{app:datasets}
%-------------------------------------------------------------------------------
In this section, we introduce two additional datasets:

\ptitle{UCI Adult dataset} We sampled 1000 data points from the UCI adult dataset \citep{dua2017} with six attributes: the three numeric attributes \emph{age}, \emph{education level}, and \emph{work hours per week}, and the three binary attributes \emph{ethnicity} (white/other), \emph{gender}, and \emph{income (>50k)}.

\dblp{
\ptitle{DBLP dataset} We extracted all papers from 20 venues\footnote{These venues are: NIPS, ICLR, ICML, AAAI, IJCAI, KDD, ECML-PKDD, ICDM, SDM, WSDM, PAKDD, VLDB, SIGMOD, ICDT, ICDE, PODS, SIGIR, WWW, CIKM, ECIR.} in four areas (ML/DM/DB/IR) of computer science from the DBLP citation network dataset \citep{tang2008arnetminer}. We sampled half of the papers and constructed a network ($122,962$ nodes\footnote{The network consists of $43,346$ paper nodes, $63,446$ author nodes, $16,150$ topic nodes and $20$ venue nodes.}) based on paper-author, paper-topic, paper-venue relations. Finally, we embedded the network into a $64$ dimensional euclidean space using node2vec \citep{grover2016node2vec} with walk length $80$, window size $10$. In our experiment, both $p$ and $q$ are set to $1$. Under this setting, node2vec is  equivalent to DeepWalk \citep{perozzi2014deepwalk}.}
%-------------------------------------------------------------------------------
\subsection{Case study: UCI Adult dataset\label{app:exp_uci_adult}}
%-------------------------------------------------------------------------------
\begin{figure*}[t]
\centering
\includegraphics[trim=0 0 0 0, clip=true, width=\textwidth]{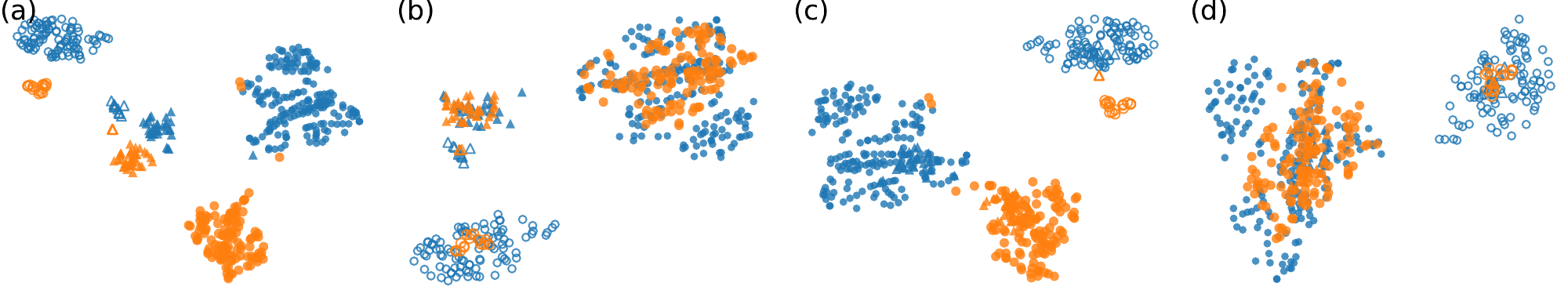}
\vskip -0.2cm
\caption{Visualization of $2$-d embeddings of the UCI Adult dataset. Points are visually encoded according to their attributes. \emph{gender}: \emph{female} (orange color), \emph{male} (blue color); \emph{ethnicity}: \emph{white} (circle), \emph{other} (triangle); \emph{income (>50k)}: \emph{true} (unfilled marker), \emph{false} (filled marker).
(a) t-SNE embedding shows clusters that are grouped according to the combinations of all three attributes. (b) With attribute \emph{gender} as prior, ct-SNE embedding shows four clusters each has a mixture of points with different genders, indicating the \emph{gender} information is removed. (c) With attribute \emph{ethnicity} as prior, ct-SNE embedding also shows four clusters but each has a mixture of points with different ethnicities. (d) Incorporating the combination of attributes \emph{gender} and \emph{ethnicity} as prior, the resulted ct-SNE embedding shows two clusters that are correlated with the remaining attribute: \emph{income (>50k)}.
\label{fig:uci_adult_emb}}
\end{figure*}

\begin{figure*}[t]
\centering
\includegraphics[trim=0 0 0 0, clip=true, width=\textwidth]{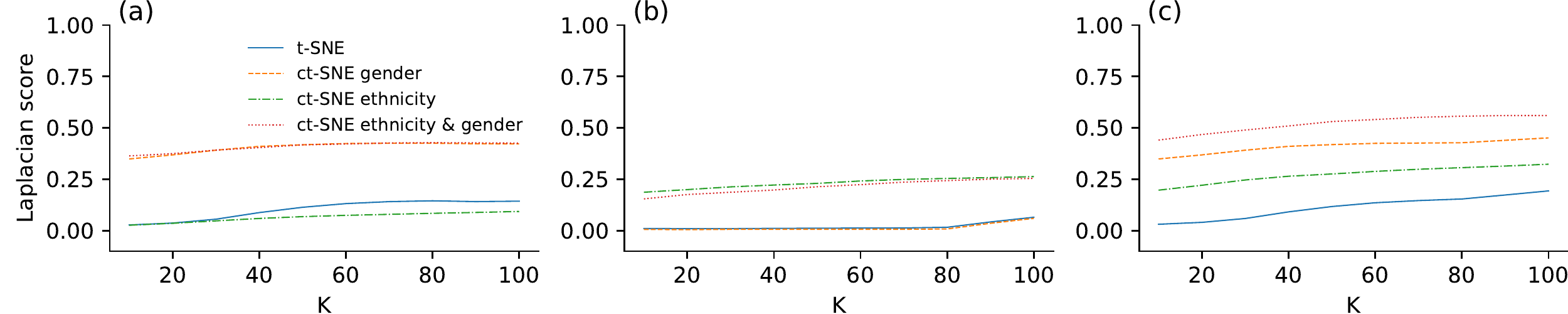}
\vskip -0.2cm
\caption{The homogeneity of cluster labels in t-SNE and several ct-SNE embeddings of the UCI Adult dataset for $k$ (a parameter of the Laplacian score) ranging from $10$ to $100$ with step size $10$. Colored lines correspond to scores for different embeddings: t-SNE (blue), ct-SNE with prior \emph{gender} (orange), ct-SNE with prior \emph{ethnicity} (green), and ct-SNE with prior \emph{ethnicity \& gender} (red). Subfigures give homogeneity scores for various labels: (a) \emph{gender} (b) \emph{ethnicity} (c) \emph{gender} \& \emph{ethnicity}.
(a) The attribute \emph{gender} has lower homogeneity (high Laplacian score) in the ct-SNE embedding with \emph{gender} or \emph{ethnicity \& gender} as prior than in t-SNE embedding and ct-SNE embedding with \emph{ethnicity} as prior. (b) The attribute \emph{ethnicity} has lower homogeneity in the ct-SNE embedding with \emph{ethnicity} or \emph{ethnicity \& gender} as priors than in the t-SNE embedding and ct-SNE with \emph{gender} as prior) embeddings.
(c) The attribute \emph{ethnicity \& gender} has high homogeneity in the t-SNE embedding only.
% (c) The attribute \emph{ethnicity\&gender} has the smallest homogeneity in ct-SNE embedding with \emph{ethnicity\&gender} labels as prior. ct-SNE embeddings with only \emph{ethnicity} or \emph{gender} as prior have stronger homogeneity against the attribute \emph{ethnicity\&gender}. The t-SNE embedding achieves the largest homogeneity in this case.
\label{fig:uci_adult_homogeneity}}
\vskip -0.4cm
\end{figure*}
\ptitle{Qualitative experiment} Fig.~\ref{fig:uci_adult_emb}a shows t-SNE gives an embedding that consists of clusters grouped according to combinations of three attributes: \emph{gender}, \emph{ethnicity} and \emph{income (>50k)}. By incorporating the attribute \emph{gender} as prior, the ct-SNE embedding (Fig.~\ref{fig:uci_adult_emb}b) contains clusters with a mixture of \emph{male} and \emph{female} points, indicating the \emph{gender} information is removed. Instead, by incorporating the attribute \emph{ethnicity} the ct-SNE embedding (Fig.~\ref{fig:uci_adult_emb}c) contains clusters with a mixture of ethnicities. Finally, incorporating the combination of attributes \emph{gender} and \emph{ethnicity} as prior, the ct-SNE embedding contains data points grouped according to \emph{income} (Fig.~\ref{fig:uci_adult_emb}d).

% \ptitle{Qualitative experiment} Figure.~\ref{fig:uci_adult_emb}a shows that t-SNE gives an embedding that consists of clusters grouped according to combinations of three attributes: \emph{gender} (coded by color), \emph{ethnicity} (coded by marker shape) and \emph{income (>50k)} (coded by filled or unfilled markers).
% By incorporating the attribute \emph{gender} as prior, the ct-SNE embedding (Fig.~\ref{fig:uci_adult_emb}b) contains four clusters where each of them contains a mixture of \emph{male} (blue) and \emph{female} (orange) points. Thus the \emph{gender} information is removed in this ct-SNE embedding.
% Supposing the attribute \emph{ethnicity} was learned first from the t-SNE embedding, and incorporating this attribute as prior, the ct-SNE embedding also contains four clusters (Fig.~\ref{fig:uci_adult_emb}c), but with each of them contains a mixture of different ethnicities (triangular and circular markers).
% Finally, incorporating the combination of attributes \emph{gender} and \emph{ethnicity} as prior, the ct-SNE embedding contains data points grouped according to \emph{income} (filled or unfilled markers, Fig.~\ref{fig:uci_adult_emb}d).

\ptitle{Quantitative experiment} We analyzed the homogeneities (Laplacian scores) of attributes \emph{gender}, \emph{ethnicity} and \emph{income (>50k)} measured on both t-SNE and ct-SNE embeddings. Fig.~\ref{fig:uci_adult_homogeneity}a shows ct-SNE with prior \emph{gender} removes the \emph{gender} factor from the resulted embedding, while ct-SNE with prior \emph{ethnicity} makes the \emph{gender} factor in the resulted embedding clearer. Similarly, Figure.~\ref{fig:uci_adult_homogeneity}b,c show ct-SNE removes the prior information effectively for labels \emph{ethnicity} and \emph{ethnicity\&gender} respectively, given the associated priors.

% \ptitle{Quantitative experiment} We analyzed the homogeneities (Laplacian score) of attributes \emph{gender}, \emph{ethnicity} and \emph{income (>50k)} measured on both t-SNE and ct-SNE embeddings. The homogeneities are measured using kNN graph with $k$ ranging from $10$ to $100$ with step size $10$. Figure~\ref{fig:uci_adult_homogeneity}a shows ct-SNE with \emph{gender} as prior (orange line) removes the \emph{gender} factor from the resulted embedding, while ct-SNE with \emph{ethnicity} prior (green line) makes the \emph{gender} factor in the resulted embedding clearer. Similarly, In Figure~\ref{fig:uci_adult_homogeneity}b, the attribute \emph{ethnicity} has small homogeneity in the ct-SNE embedding that has \emph{ethnicity} as prior. The ct-SNE embedding with \emph{gender} as prior gives attribute \emph{ethnicity} stronger homogeneity. Finally, ct-SNE embedding with prior that combines attribute \emph{ethnicity} and \emph{gender} (red line) clearly reveals  the \emph{income (>50k)} attribute (Fig.~\ref{fig:uci_adult_homogeneity}c).

%-------------------------------------------------------------------------------
\subsection{Case study: DBLP dataset}\label{app:exp_dblp}
%-------------------------------------------------------------------------------
\begin{figure*}[tp]
\centering
\includegraphics[width=0.8\textwidth]{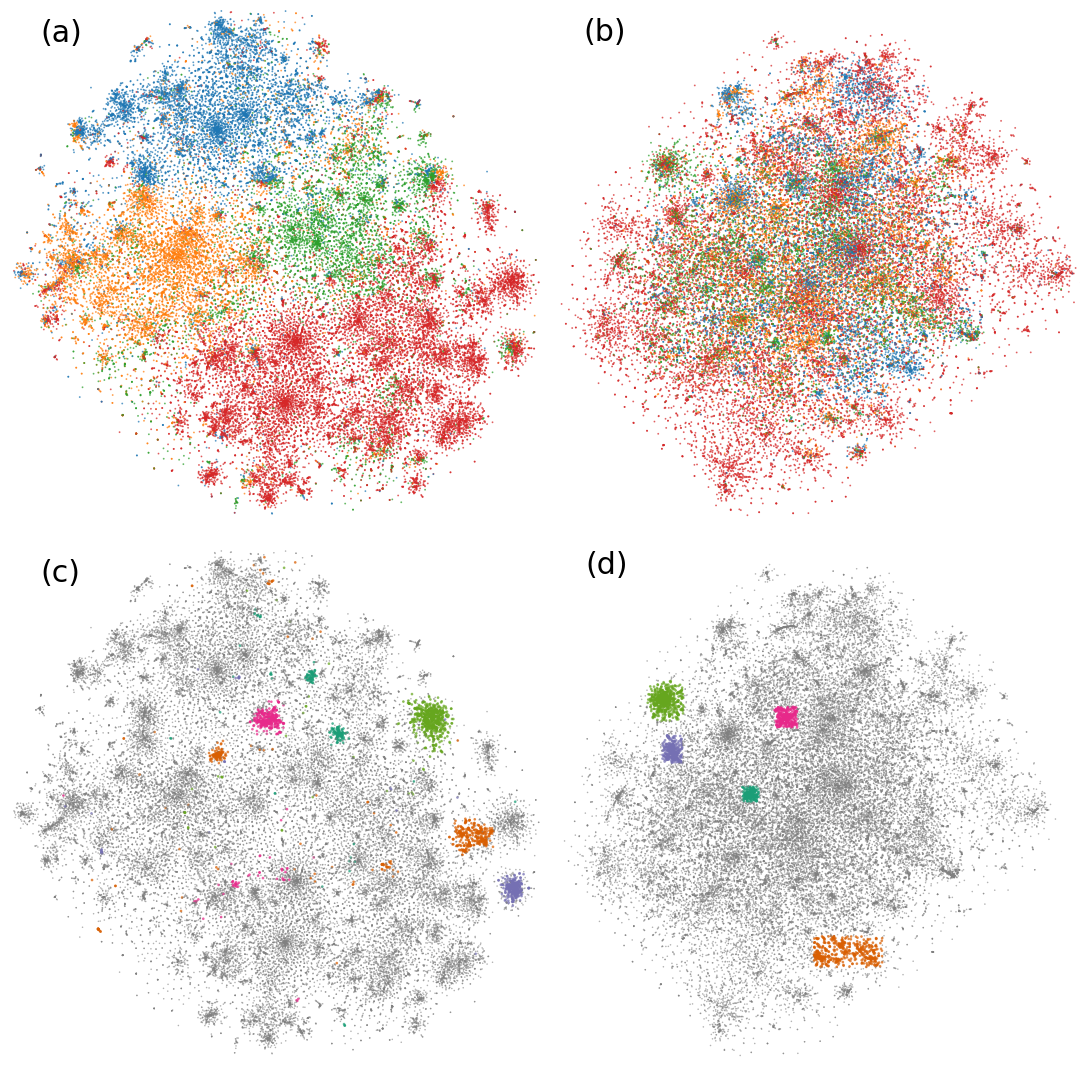}
\vskip -0.4cm
\caption{\dblp{Visualization of 2-d embeddings of the DBLP dataset. Left column: t-SNE embedding, right column: ct-SNE embedding with area as prior. The rows contains different cluster markings. (a) t-SNE embedding shows a clustering according to four areas in computer science (red - machine learning, green - data mining, blue - data base, orange - information retrieval). (b) ct-SNE embedding shows a different clustering, with area information removed. (d) Newly emerged visual clusters (magenta - topic `privacy', dark green - topic `data stream', orange - topic `computer vision') in ct-SNE embedding spread over in the t-SNE embedding (c), corresponding to users interested in horse riding. (d) Clusters (grass green - topic `clustering', purple - topic `active leraning') stood-out in the ct-SNE embedding also exists in the t-SNE embedding (c). These are a few out of many clusters that we found to exhibit a much more informative, interest-centric structure than the t-SNE projection.} \label{fig:dblp_emb}}
\vskip -0.4cm
\end{figure*}
\ptitle{Qualitative experiment}
\dblp{Applying t-SNE on the DBLP dataset gives a visualization with many visual clusters (Fig.~\ref{fig:dblp_emb}a). Feature ranking for classification of the selected clusters shows the topics that contribute the most to the visualization. Moreover, we used mpld3\footnote{\url{https://mpld3.github.io}} (an interactive visualization library) to inspect (i.e., hovering over data points and check tooltips) the metadata of t-SNE plot. Upon inspection, the visualization appears to be globally divided according the four areas. This is further confirmed by coloring the data points according to the four areas: most of the clusters are indeed quite homogeneous with respect areas}

\dblp{Knowing from the t-SNE visualization the papers are indeed divided according to areas, the area structure in the visualization is not very informative anymore. Thus we can encode the area as prior for ct-SNE so that other interesting structures can emerge. Using the same color scheme, ct-SNE shows a visualization that has many clusters with mixed colors (Fig.~\ref{fig:dblp_emb}b). This indicates the area information is mostly removed in the ct-SNE embedding. This is further confirmed by selecting clusters in ct-SNE embedding (Fig.~\ref{fig:dblp_emb}d) and highlight the same set of points in the t-SNE embedding (Fig.~\ref{fig:dblp_emb}c). The clusters highlighted in the ct-SNE visualization often consists of clusters (topics) from different areas (i.e., t-SNE clusters with different colors) that spread over the t-SNE visualization. Indeed, feature ranking indicates that papers in the selected ct-SNE cluster have similar topics in e.g., `privacy', `data steam', `computer vision'. Finally, we noticed that some clusters in ct-SNE (Fig.~\ref{fig:dblp_emb}d) embedding also exist in the t-SNE embedding (Fig.~\ref{fig:dblp_emb}c). Using feature ranking as above we found these clusters are not homogeneous in terms of area of study, but in terms of topics (e.g., `clustering', `active learning'), indicating a tightly connected research community behind the topic. Thus, by removing the irrelevant area structure using ct-SNE, clusters that persists in both visualizations become more salient and easier to observe.}
%-------------------------------------------------------------------------------
\subsection{Parameters sensitivity}\label{app:exp_param}
%-------------------------------------------------------------------------------
\begin{figure*}[t]
\centering
\includegraphics[trim=0 0 0 0, clip=true, width=\textwidth]{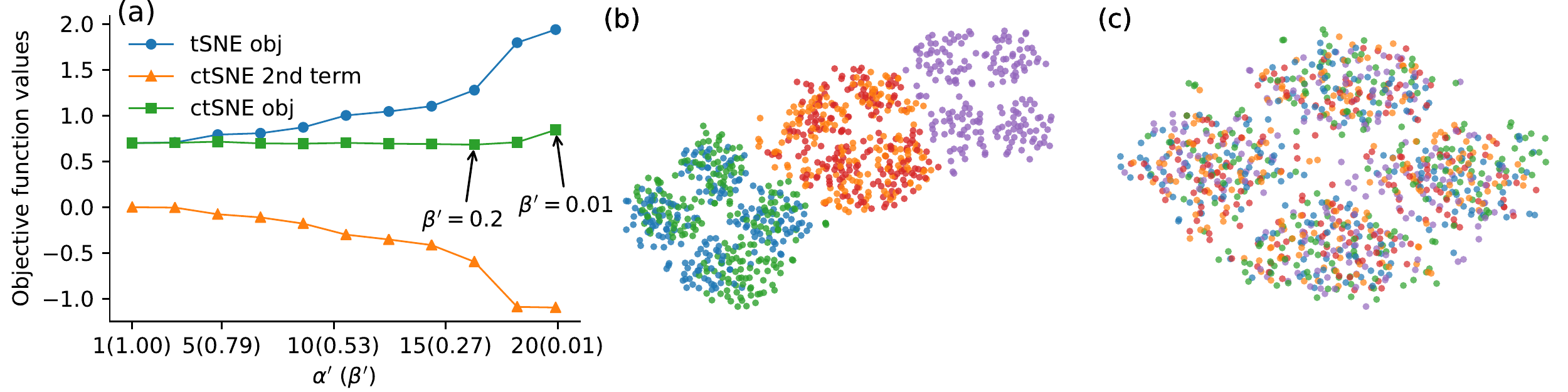}
\caption{Visualizing the effect of different $\beta'$s ($\alpha'$s) have on the ct-SNE embeddings. The embeddings are computed on the synthetic dataset with the prior information to be the cluster labels in dimensions $1$-$4$.
(a) The values of ct-SNE objective (green), t-SNE objective (blue), and ct-SNE prior term (orange) against different $\beta'$s. ct-SNE achieves smaller KL-divergence than t-SNE. (b) ct-SNE embedding with $\beta' = 0.2$ has smallest KL-divergences but is not the best visualization. (c) ct-SNE embedding with $\beta'=0.01$ gives a better visualization.
\label{fig:parameter_sensitivity}}
\vskip -0.3cm
\end{figure*}
To understand the effect of the parameter $\alpha'$ (or equivalently, $\beta'$) on ct-SNE embeddings (\textbf{Q3}), we study ct-SNE embeddings on the synthetic dataset with the prior fixed to be the cluster labels in dimensions 1--4. First, we try to understand the relation between the ct-SNE objective and the parameter $\alpha'$ (or equivalently, $\beta'$). We evaluated the ct-SNE objective (Eq.~\ref{eq:ct-SNE_obj}) on the ct-SNE embeddings obtained by ranging $\beta'$ (and $\alpha'$ correspondingly) from $0.01$ (strong prior removal effect) to $1.0$ (no prior remove effect, equivalent to t-SNE) with step size $0.1$. We also evaluated the t-SNE objective (first term in Eq.~\ref{eq:ct-SNE_obj}) and the second term in Eq.~\ref{eq:ct-SNE_obj} (the only term that depends on the prior, subsequently referred to as the \emph{prior term}) for the ct-SNE embeddings associated with various $\beta'$s.

Fig.~\ref{fig:parameter_sensitivity}a visualizes the values of the ct-SNE objective, t-SNE objective, and ct-SNE prior term against different $\beta'$s. Observe that by using a prior, the ct-SNE embedding achieves a better approximation to the higher dimensional data. That is, ct-SNE achieves a lower KL-divergence (lowest at $\beta' = 0.3$) than t-SNE does ($\beta' = 1$). This is because the prior term in the ct-SNE objective can be negative. Although the t-SNE objective increases when $\beta'$ decreases, it is compensated by the negative value contributed by the prior term. Indeed, by factoring out certain prior from the lower dimensional embedding, the necessity of the embedding to represent the prior is alleviated, enabling ct-SNE to have more freedom to approximate the high-dimensional proximities.

Interestingly, we observe that the embedding with smallest KL-divergence does not necessarily give better visualization (e.g., clear separation of the clusters). We visualize the ct-SNE embedding that achieves smallest KL-divergence ($\beta'=0.3$, Fig.~\ref{fig:parameter_sensitivity}b) and compare it with the ct-SNE embedding that has strongest prior removal effect but larger KL-divergence ($\beta'=0.01$, Fig.~\ref{fig:parameter_sensitivity}c). Although the embedding with stronger prior removal effect has larger objective value, it gives a clearer clustering than in the embedding with smaller KL-divergence ($\beta'=0.3$). As a result, the clusters in dimensions 5--6 are easier to identify. Hence, we propose as rule of thumb when using ct-SNE for visualization to use small $\beta'$ (e.g., $\beta'=0.01$).

%-------------------------------------------------------------------------------
\subsection{Baseline comparisons}\label{app:exp_baseline}
%-------------------------------------------------------------------------------
In this section, we compare ct-SNE with two non-trivial baselines. The basic idea is to first remove the known factor from the dataset, and perform t-SNE to produce lower dimensional representations. Here we use a non-linear and a linear method to remove the known factors: adversarial auto-encoder (AAE) and canonical correlation analysis (CCA). The implementation of the baselines and code for comparison experiments are also available at \url{https://bitbucket.org/ghentdatascience/ct-sne}.

\ptitlenoskip{Baseline: AAE and t-SNE} Adversarial auto-encoder (AAE) \cite{makhzani2015adversarial} can be used to learn a latent representation that prevents the discriminator from predicting certain attributes \citep{madras2018learning}. In order to remove prior information from the low-dimensional representation of a dataset using AAE, we can configure the discriminator to predict the prior attributes, and using the auto-encoder to adversarially remove the prior from the latent representation of the dataset.

\begin{figure*}[t]
\centering
   \includegraphics[trim=0 0 0 0, clip=true, width=\textwidth]{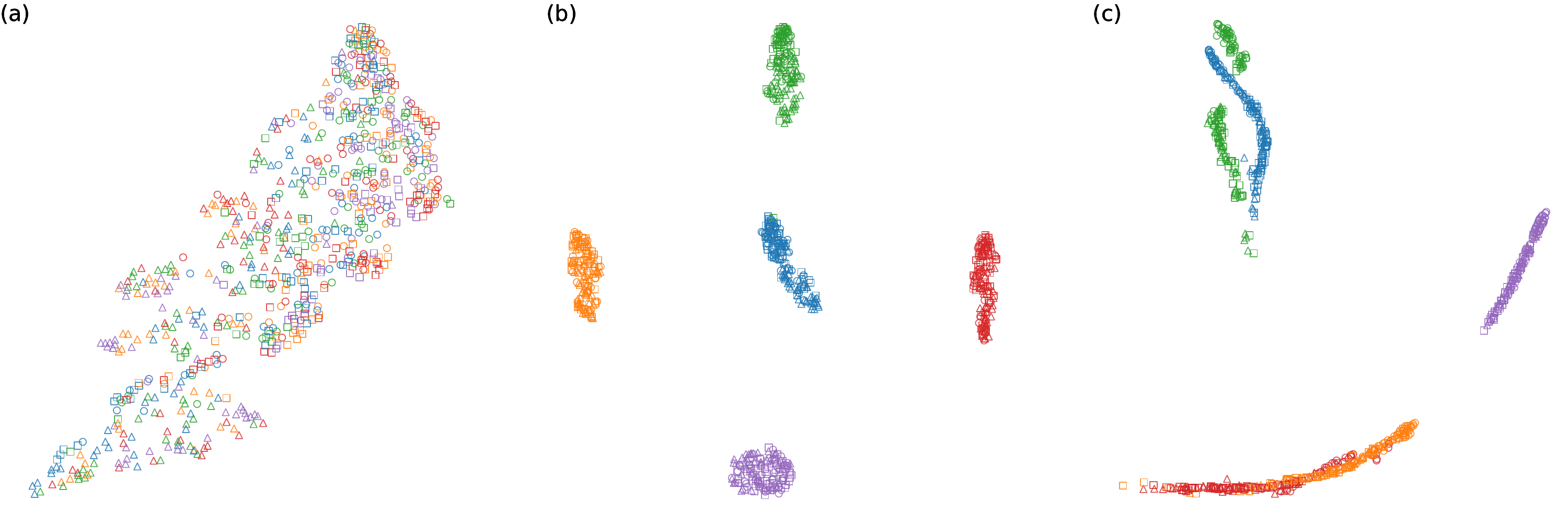}
\caption{
Visualization of 2-d embeddings obtained by applying the AAE based approach on the synthetic dataset. The data points are colored according to the cluster label in dimensions $1$-$4$. The data points are also plotted using different markers based on the cluster labels in dimensions $5$-$6$. (a) The AAE based approach successfully removed the clustering information in dimensions $1$-$4$, but failed to reveal the clusters in dimensions $5$-$6$ (b) AAE successfully removed the clustering information in dimensions $5$-$6$ and also reveals the clusters in dimensions $1$-$4$ (c) AAE failed to remove the clustering information in dimensions $1$-$6$. \label{fig:baseline_aae}}
\end{figure*}

We adopt the AAE configuration described by \citet{edwards2015censoring}. AAE is in general difficult to tune: it has $8$ hyperparameters ($4$ network structure parameters, $2$ weights in the objective, and $2$ learning rates) and a few design choices about the network architecture (e.g., the number of layers in each subnetwork and activation functions). We tried different parameter settings and managed to remove the clustering label information in dimensions 1--4 (Fig.~\ref{fig:baseline_aae}a) and 5--6 (Fig.~\ref{fig:baseline_aae}b) from the data. In Figure~\ref{fig:baseline_aae}a, the AAE approach manages to remove the prior information, but it fails to pick up the complementary structure in the data (clusters in dimensions 5--6). It also fails to remove the prior information (cluster labels in dimension 1--6) in Figure~\ref{fig:baseline_aae}c. Comparing to this baseline, ct-SNE practically has only one parameter ($\beta'$) to tune, which often can be set to a small positive number (e.g., $0.01$).

\begin{figure*}[t]
\centering
   \includegraphics[width=\textwidth]{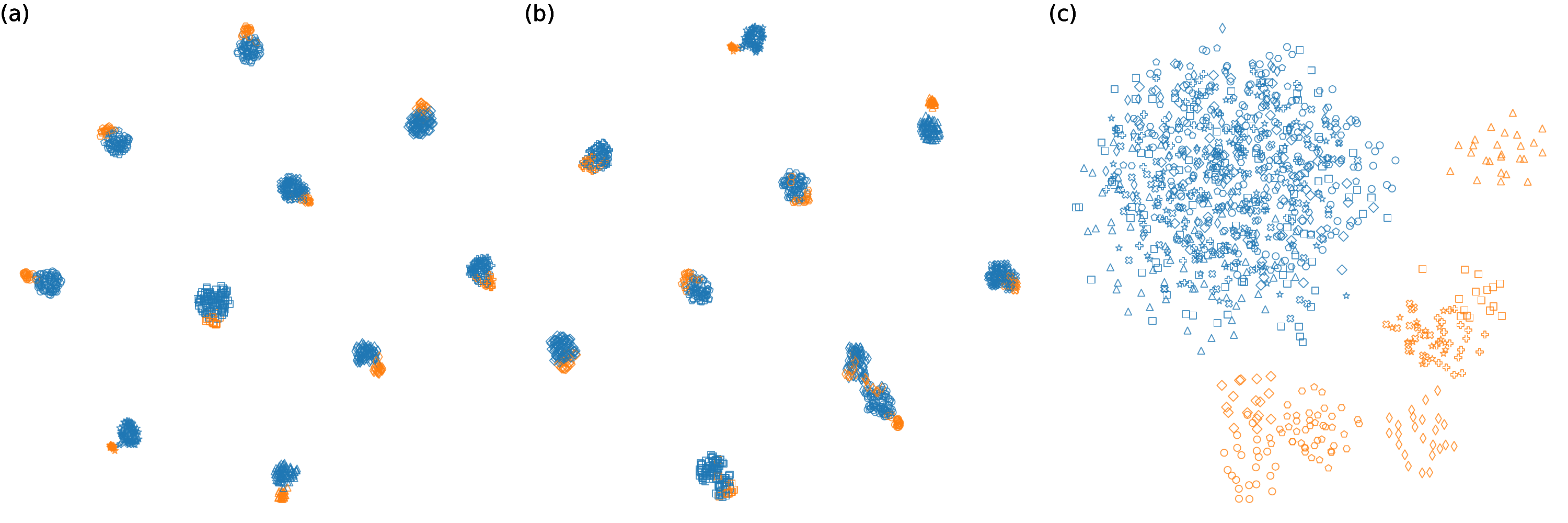}
\caption{Visualization of 2-d embeddings obtained by applying CCA-based approaches and ct-SNE on a synthetic $5$ dimensional dataset. (a) Projecting data onto the null space of CCA top components and then apply t-SNE gives an embedding that picks up the $10$ large clusters (plotted with different markers) but failed to pick up the structure of two small clusters (colored differently) within each large cluster. (b) Projecting the data onto CCA components with least correlation and then apply t-SNE also fails to pick up the two-cluster structure within the large clusters. (c) ct-SNE removes the $10$ cluster information in the embedding and shows clearly the two cluster structure within each larger cluster. \label{fig:baseline_cca}}
\end{figure*}

\ptitle{Baseline: CCA and t-SNE}
Canonical correlation analysis \citep{hotelling1936relations} aims to find a  linear transformation for two random variables such that the correlation between transformed variables is maximized. To remove the prior information from data using CCA, one approach is to first find the (at most) $d-2$ subspace ($d$ is the dimensionality of the data) in which the transformed data and the prior information (one hot encoding of the labels) have the largest correlation. Then the data is whitened by projecting it onto the null space (at least $2$-d) of the subspace found in the first step. By doing so, the whitened data is less correlated to the known factor.

Another variant of the CCA-based approach is directly projecting the data onto the $2$-dimensional subspace found by CCA in which the transformed data and labels has smallest correlation. To be consistent, we also apply t-SNE to the transformed data.

Our experimental results show the CCA-based approaches can easily remove label information that is orthogonal to other attributes in the data. For example, in the UCI Adult dataset, the gender information is orthogonal to the ethnicity and income, which can be easily removed using the CCA approach. However, the CCA-based approach performs poorly when the known factor is correlated with other attributes. Moreover, the CCA-based approaches also have the limitation that the number of the projection vectors is upper-bounded by the dimensionality of the data. If the number of unique values of an attribute exceeds the dimensionality of the data, the CCA projection would not be able to remove the label info entirely from the data. To illustrate our points, we synthesized a $5$-dimensional dataset with 1,000 data points. The data points are grouped into $10$ clusters each corresponding to a multi-variate Gaussian with random location and small variance. Additionally, each cluster is separated into two small clusters (one contains $20\%$ points of the cluster, and another includes the rest) along one randomly chosen axis. Figure~\ref{fig:baseline_cca}a,b shows both the CCA approaches pick up only the $10$ large clusters (differentiated using marker shape) but failed to pick up the structure of two small clusters (plotted in different colors) within each large cluster. On the other hand, ct-SNE removes the $10$ cluster information in the embedding and shows each large cluster can be further separated in to two smaller clusters.

Thus, the CCA-based baselines perform poorly when the known factor is correlated with other attributes. Moreover, the number of the projection vectors in CCA-based baselines is upper-bounded by the dimensionality of the data. Meanwhile, ct-SNE does not have these limitations.

%-------------------------------------------------------------------------------
\subsection{Runtime}\label{app:exp_runtime}
%-------------------------------------------------------------------------------
We measure the runtime of the exact ct-SNE and the approximated version ($\theta=0.5$) on a PC with a quad-core $2.3$GHz Inter Core i5 and a 2133MHz LPDDR3 RAM. By default, the maximum number of iterations of ct-SNE gradient update is 1,000. For larger datasets and prior attributes that have many values, more iterations are required to achieve a convergence. For example, the synthetic dataset (1,000 samples and 10 dimensions) requires fewer than 1,000 iterations to converge while the \fb{Facebook dataset (500,000 examples and 128 dimensions) requires 3,000 iterations to converge.} Table.~\ref{tab:runtime} shows that approximated ct-SNE is efficient and applicable to large data with high dimensionality, while exact ct-SNE is not.
\begin{table}[tb]
  \centering
  \caption{\label{tab:runtime} Average runtime (in seconds) of exact and approximated ct-SNE in computing one gradient update step. To measure the runtime of ct-SNE on a dataset with similar size as the Facebook dataset, we scaled the Synthetic dataset up to $500,000$ data points with $128$ dimensions.}
  \vskip -0.2cm
  \begin{tabular}{c | c | c | c | c }
    name & size & dim. & exact & apprx. ($\theta = 0.5$) \\\hline
    Synthetic & 1,000 & 10 & 0.06& 0.01\\
    UCI Adult & 1,000 & 6 & 0.07 & 0.01 \\
    DBLP & 43,346 & 64 & 503.97 & 0.45\\
    Synthetic & 500,000 & 128 & 100,278 & 9.1 \\
  \end{tabular}
  \vskip -0.5cm
\end{table}

%===============================================================================
\section{Extended related work\label{app:related_work}}
%===============================================================================
Many dimensionality reduction methods have been proposed in the literature. Arguably, $n$-body problem based methods\footnote{In Section~\ref{sec:optimization} we provide more information on the $n$-body problem} such as MDS \citep{torgerson1952multidimensional}, Isomap \citep{tenenbaum2000global}, t-SNE \citep{maaten2008visualizing}, LargeVis \citep{tang2016visualizing}, and UMAP \citep{mcinnes2018umap} appear to be the most popular ones. These methods typically have three components: (1) a proximity measure in the input space, (2) a proximity measure in the embedding space, (3) a loss function comparing the proximity between data points in the embedding space with the proximity in the input space. When minimizing the loss over the embedding space, the data points (i.e., the $n$ bodies) have pairwise interactions and the embedding of all points needs to be updated simultaneously. Since the optimization problem is not convex, local minima are typically accepted as output. ct-SNE belongs to this class of DR methods. It accepts both high-dimensional data and priors about the data as inputs, and searches for low-dimensional embeddings while discounting structure in the input data specified as prior knowledge. Closely related, in the multi-maps t-SNE work \citep{van2012visualizing} factors that are mutually exclusive are captured by multiple t-SNE embeddings at once. Comparing to multi-map t-SNE, ct-SNE allows users to disentangle information in a targeted (subjective) manner, by specifying which information they would like to have factored out.

As a core component of ct-SNE is the prior information specified by the user, it can be considered an interactive DR method. Existing papers on \emph{interactive} DR can be categorized into two groups. The first group aim to improve the explainability and computation efficiency of existing DR methods via novel visualizations and interactions. iPCA \citep{jeong2009ipca} allows users to easily explore the PCA components and thus achieve better understanding of the linear projections of the data onto different PCA components. \citet{cavallo2018visual} helps the user to understand low-dimensional representations by applying perturbations to probe the connection between input attributed space and embedding space. Similarly, \citet{faust2019} introduce a method based on perturbations to visualize the effect of a specific input attribute on the embedding, while \citet{stahnke2016} introduce `probing' as a means to understand the meaning of point set selections within the embedding. Steerable t-SNE \citep{pezzotti2017approximated} aims to make t-SNE more scalable by quickly providing a sketch of an embedding which is then refined only upon the user's interests.
% and refining the embedding only in the region of the user's interests.

The second group of interactive DR methods adjust the algorithms according to a users' inputs. SICA \citep{kang2016subjectively} and SIDE \citep{puolamaki2018interactive} explicitly model the user's belief state and find linear projections that contrast to it. These two methods are linear DR methods thus cannot present non-linear structures in the low-dimensional representations. Work by \citet{diaz2014interactive} allows users to define their own metric in the input space, after which the low-dimensional representation reflects the adjusted importance of the attributes. This method puts the burden on the user for direct manipulation of the input space metric. Many variants of existing DR methods have been introduced where user feedback entails editing of the embedding, and such manually embedded points are used as constraints to guide the dimensionality reduction \citep[e.g.,][]{alipanahi2011,barshan2011,paurat2013}. These methods contrast with ct-SNE in that the user feedback must be obeyed in the output embedding, while for ct-SNE the prior knowledge defined by the user guides what is irrelevant to the user.

\end{document}

%% file: math_commands.tex
\usepackage{amsmath,amsfonts,bm}

\def\eqref#1{equation~\ref{#1}}
\def\1{\bm{1}}

\def\vf{{\bm{f}}}

\def\vp{{\bm{p}}}
\def\vq{{\bm{q}}}
\def\vr{{\bm{r}}}

\def\vx{{\bm{x}}}
\def\vy{{\bm{y}}}

\def\mA{{\bm{A}}}

\def\mD{{\bm{D}}}

\def\mL{{\bm{L}}}

\def\mX{{\bm{X}}}
\def\mY{{\bm{Y}}}

\DeclareMathAlphabet{\mathsfit}{\encodingdefault}{\sfdefault}{m}{sl}
\SetMathAlphabet{\mathsfit}{bold}{\encodingdefault}{\sfdefault}{bx}{n}

\def\sR{{\mathbb{R}}}